\documentclass[letterpaper, 10 pt, conference]{ieeeconf}  % Comment this line out if you need a4paper

\IEEEoverridecommandlockouts                              % This command is only needed if 
\usepackage{graphics}
\usepackage{epsfig} % for postscript graphics files
\usepackage{mathptmx} % assumes new font selection scheme installed
\usepackage{times} % assumes new font selection scheme installed
\usepackage{amsmath} % assumes amsmath package installed
\usepackage{amssymb}  % assumes amsmath package installed
\usepackage{multirow}
\usepackage{url}
\usepackage{caption}
\usepackage{subcaption}
\usepackage[table]{xcolor}
\usepackage{import}
\usepackage{hyperref}
\usepackage{cleveref}
\usepackage{comment}
\usepackage{booktabs}
\usepackage{tabularx}
\usepackage{gensymb} % For \degree command when using also siunitx package
\usepackage{cite}
\usepackage{stfloats}
\usepackage{bm}
\usepackage{siunitx}
\usepackage{acro}
\DeclareAcronym{com}{ 
    short = {CoM}, 
    long  = {Center of Mass},
    first-style = short-long,
}
\DeclareAcronym{cog}{ 
    short = {CoG}, 
    long  = {Center of Geometry},
    first-style = short-long,
}

\DeclareAcronym{wrt}{ 
    short = {w.r.t.}, 
    long  = {with respect to},
    first-style = short-long,
}
\DeclareAcronym{ee}{ 
    short = {EE}, 
    long  = {End-Effector},
    first-style = short-long,
}
\DeclareAcronym{dof}{ 
    short = {DoF}, 
    long  = {Degree of Freedom},
    first-style = short-long,
}
\usepackage{csquotes}
\newcommand\new[1]{\textcolor{black}{#1}}

\title{\LARGE \bf Advancing Manipulation Capabilities of a UAV Featuring Dynamic Center-of-Mass Displacement}

\author{ Tong Hui$^*$, Matteo Fumagalli
\thanks{This work has been supported by the European Unions Horizon 2020 Research and Innovation Programme AERO-TRAIN under Grant Agreement No. 953454. $^*$ Corresponding author email: {\tt\small tonhu@dtu.dk}}% <-this % stops a space
\thanks{The authors are with the Technical University of Denmark, Denmark.}}

\begin{document}

\maketitle
\thispagestyle{empty}
\pagestyle{empty}

%%%%%%%%%%%%%%%%%%%%%%%%%%%%%%%%%%%%%%%%%%%%%%%%%%%%%%%%%%%%%%%%%%%%%%%%%%%%%%%%
\begin{abstract}
As aerial robots gain traction in industrial applications, there is growing interest in enhancing their physical interaction capabilities. Pushing tasks performed by aerial manipulators have been successfully demonstrated in contact-based inspections. However, more complex industrial applications require these systems to support higher-\ac{dof} manipulators and generate larger forces while pushing (e.g., drilling, grinding). 
This paper builds on our previous work, where we introduced an aerial vehicle that can dynamically vary its \ac{com} location to improve force exertion during interactions. We propose a novel approach to further enhance this system’s force generation by optimizing its \ac{com} location during interactions. Additionally, we study the case of this aerial vehicle equipped with a 2-\ac{dof} manipulation arm to extend the system's functionality in tool-based tasks. The effectiveness of the proposed methods is validated through simulations, demonstrating the potential of this system for advanced aerial manipulation in practical settings.
% Aerial manipulators can serve contact-based industrial applications, where fundamental tasks like drilling and grinding often necessitate aerial platforms to handle heavy tools and high loads (i.e., forces and torques). These tasks are frequently performed on non-horizontal surfaces. Current multirotor platforms, which have a fixed \ac{com} within the rotor-defined area, typically exhibit a large moment arm between the \ac{ee} tip and the system's \ac{com}. This configuration can result in instability and potential damage during physical interactions.
% In this letter, we present the system design, modeling, and control of a novel aerial vehicle tailored to tool manipulation on non-horizontal surfaces. This platform adapts the form of an H-shaped coaxial octocopter with tiltable back rotors; it can carry heavy components (e.g., the manipulator and battery) on a movable plate within the rotor-defined area during free flight. When interacting with surfaces, the platform actively shifts the plate toward the work surface while preserving the system orientation thanks to the tiltable back rotors. This leads to a displaced \ac{com} location and a reduced moment arm. We use simulations that closely capture the built physical prototype to validate our proposed concepts for complex and risky working scenarios. Moreover, early-stage physical experiments were conducted to evaluate the developed system for free flights and a pushing task.
\end{abstract}

%%%%%%%%%%%%%%%%%%%%%%%%%%%%%%%%%%%%%%%%%%%%%%%%%%%%%%%%%%%%%%%%%%%%%%%%%%%%%%%%
\section{Introduction} \label{sec:intro}
With the growing interest in using aerial robots for industrial applications, substantial research has been dedicated to developing aerial manipulation systems \cite{review}. The aerial robotics community is exploring various physical interaction tasks to expand their industrial use cases \cite{tong_iser}. While aerial manipulators have proven effective in contact-based inspections through pushing tasks \cite{bart,truj,bodie2019,watson2022,tong_icra_1,tong_icra_2}, advanced tool manipulation for broader industrial applications often requires platforms to carry higher-\ac{dof} manipulators and exert larger forces on work surfaces \cite{arm,sun,drill,ding}.

Common 4-\ac{dof} multirotors often have limited force generation capabilities in both direction and magnitude when interacting with the environment \cite{review,tong_icra_1}. To overcome this, thrust vectoring \cite{truj,antonio,odar,mina,watson2022,ding,hwang,bodie2019} and perching technologies \cite{meng,drill,backus,zhang,daler,wop} have been applied to enhance force generation in aerial systems. Notably, most of the existing platforms feature a fixed system's \ac{com} position.
\begin{figure}[!t]
   \centering
   %\vspace{0.25cm}
\includegraphics[width=\columnwidth]{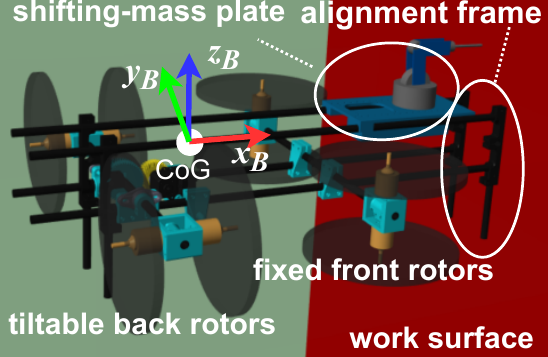}
    \caption{A still frame from the tool manipulation simulation of the designed aerial vehicle. The body frame $\mathcal{F}^B=\{O_B;\bm{x}_B,\bm{y}_B,\bm{z}_B\}$ is attached to the aerial vehicle \ac{cog}. The aerial vehicle preserves contact with a vertical surface using an alignment frame along the body axis $\bm{x}_B$ supported by tilted back rotors. The system allows a dynamic \ac{com} displacement by moving the shifting-mass plate toward the work surface while preserving the orientation and contact using the alignment frame. A 2-\ac{dof} manipulator arm is mounted on the shifting-mass plate which allows tool operation closer to the work surface while displacing the \ac{com}.}
    \label{fig:physical_model}
    \vspace{-0.5cm}
\end{figure}

Perching is commonly used to handle heavy tools and exert significant forces on non-horizontal surfaces \cite{meng}. While grasping-based perching adds complexity and weight due to grippers \cite{backus,zhang}, attachment-based perching using magnets, adhesives, or vacuum cups has gained popularity \cite{meng,daler,wop,drill}. In \cite{drill}, a perching and tilting aerial vehicle for concrete wall drilling was introduced, using two suction cups to perch and bring its fixed \ac{com} closer to the work surface. However, the perching success depends heavily on surface conditions like flatness, roughness, shape, humidity, and other environmental factors, which can compromise robustness \cite{meng}.

In \cite{mina,truj,antonio,odar, watson2022,bodie2019}, authors present aerial vehicles achieving 6-\ac{dof} wrench generation in the space being fully-actuated.
These platforms however often suffer from counteracting unwanted thrust components, high energy consumption, and control complexity.
Alternatively, some solutions, such as H-shaped aerial vehicles \cite{ding,hwang}, employ 5-\ac{dof} actuation by tilting both front and back rotors. While effective, similar tasks can be accomplished by tilting only one set of rotors (either front or back), reducing actuation complexity. However, this minimalist rotor configuration poses challenges when large forces need to be exerted on the environment.

In our previous work \cite{ral,tmech}\footnote{The associated video is available at \url{https://youtu.be/NXEnCS1ZLEg}.}, we introduced a novel perspective to address the force generation problem in aerial manipulation by considering varied \ac{com} locations. We proposed an innovative aerial system design with a dynamically displacing \ac{com} configuration to enhance force generation during interactions supported by tilting only back rotors. This system allows smooth transitions between free flight and high-force interactions.

\subsection{Contribution}
In this work, we aim to further improve the manipulation capabilities of our previously developed aerial system. We propose a method to optimize the system's \ac{com} location during interactions in the presence of modeling inaccuracy and friction effects at contact, fully utilizing its dynamic \ac{com} design to enhance the force exertion of the system. Additionally, we study the case of tool manipulation using a 2-\ac{dof} arm attached to the aerial vehicle (see Fig.~\ref{fig:physical_model}). We provide an in-depth analysis of the potential operational flow and practical limitations of such tool manipulation tasks. The proposed concepts are validated through realistic simulations, demonstrating the feasibility and advantages of this system for advanced aerial manipulation applications.

\section{Background}\label{sec:background}
\begin{figure}[!t]
   \centering
\begin{subfigure}{0.44\columnwidth}\centering
    \includegraphics[trim={0.8cm 0.4cm 0.8cm  0.0cm},clip,width=\columnwidth]{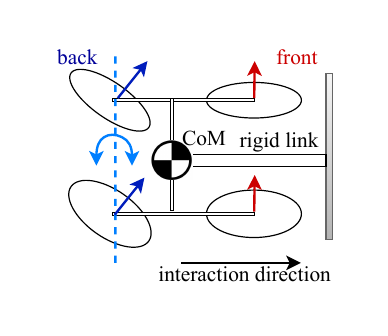}
    \caption{}
    \label{fig:design1}
\end{subfigure}
     \begin{subfigure}{0.54\columnwidth}\centering
    \includegraphics[trim={0.8cm 0.5cm 0.7cm  0.5cm},clip,width=\columnwidth]{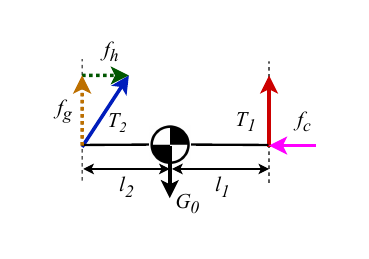}
    \caption{}
    \label{fig:design2}
\end{subfigure}
\caption{(a) A simplified illustration of an H-shaped aerial vehicle with tiltable back rotors and fixed front rotors pushing on a vertical surface using the attached rigid link. (b) Free body diagram of the system in (a) \cite{ral}.}\label{fig:design_motivation}
    %\vspace{-0.5cm}
\end{figure}

In our previous work \cite{ral}, we studied how varying the \ac{com} location affects force exertion in aerial manipulation, motivating the design of an aerial vehicle with a dynamically displacing \ac{com}. Specifically, we analyzed the performance of an H-shaped aerial system with tilting back rotors and fixed front rotors, as shown in Fig.~\ref{fig:design1}. The system features a rigid link for interacting with a vertical surface during a pushing task. 

Our analysis focuses on how the \ac{com} position influences the system’s ability to exert pushing forces on vertical surfaces, rather than to precisely quantify exerted forces. Friction forces at contact points are omitted in this analysis, as they do not significantly impact the overarching results. In practical implementations, such friction effects are treated as external disturbances, consistent with standard assumptions in aerial robot pushing operations, and are excluded from the control design framework.
\begin{figure}[!t]
   \centering
   %\vspace{0.25cm}
\includegraphics[trim={0.8cm 0.5cm 0.7cm  0.5cm},clip,width=0.7\columnwidth]{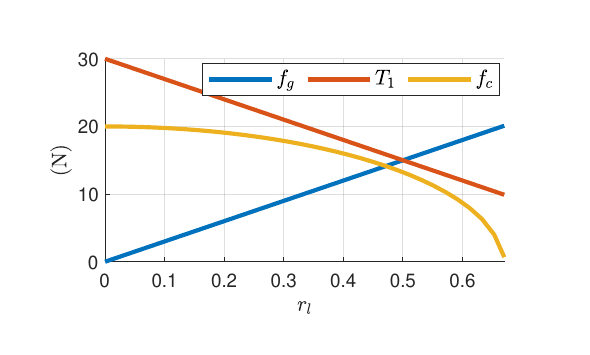}
    \caption{The variation of forces $f_c$, $T_1$, and $f_g$ \ac{wrt} $r_l$ based on \cref{eq:force_com,eq:force_com2,eq:force_com3} \cite{ral}.}
    \label{fig:force_com}
    %\vspace{-0.5cm}
\end{figure}
\subsection{Wrench Analysis}
We consider symmetry around the rigid link’s main axis and simplify the problem to a planar scenario (Fig.\ref{fig:design2}). Denoting $T_1$ and $T_2$ as the thrust of the front and back rotors and $G_0$ as the gravitational force, we define $f_h$ as the component of $T_2$ along the interaction direction and $f_g$ as the component opposing gravity. The contact force from the environment, passing through the system's \ac{com}, is $f_c$, and $l_1$, $l_2$ represent the distances between the propellers and the \ac{com}. Assuming equilibrium during interaction, the forces follow these relationships:
\begin{align}\label{eq:force_com}
    f_g&=\frac{G_0l_1}{l_1+l_2}=G_0r_l,\\
    T_1&=G_0(1-r_l)\label{eq:force_com2},\\
    f_c&=f_h=\sqrt{T_2^2-f_g^2}\label{eq:force_com3}.
\end{align}
where $r_l=\tfrac{l_1}{l_1+l_2}$ and $l_1+l_2$ is fixed. We analyzed the system with $G_0=30\si{\newton}$, $T_2=20\si{\newton}$, and $l_1+l_2=0.27\si{\meter}$, varying $r_l$ and calculating the force variations (Fig.~\ref{fig:force_com}).

\subsection{Optimal CoM Location}
\label{sec:optimal}
In traditional designs with a fixed \ac{com} ($r_l=0.5$), $f_h=f_c$ is limited to $0.65T_2$, see Fig.~\ref{fig:force_com}. Shifting the \ac{com} closer to the back rotors ($r_l>0.5$) reduces force exertion, with $f_c$ approaching zero at $r_l=0.7$. Conversely, moving the \ac{com} closer to the front rotors ($r_l<0.5$) enhances force generation, with maximum exertion ($f_c=T_2$) when $r_l=0$. In our context, we refer to the \ac{com} position at $r_l=0$ as the optimal \ac{com} location for enlarging force generation during interactions. The consideration of friction effects on optimal \ac{com} location is discussed later in Sec.~\ref{sec:problem}.
\subsection{A Novel Aerial Vehicle Design}
\label{sec:design_concept}
This analysis motivates a novel approach to address high-force interaction in aerial manipulation: shifting the system's \ac{com} closer to the front rotors enhances force exertion while pushing. We proposed an innovative aerial system that operates as a 4-\ac{dof} aerial vehicle with a fixed \ac{com} during navigation being energy-efficient. This aerial vehicle enables dynamic \ac{com} displacement and tiltable back rotors upon execution of a pushing task. Such design increases the system's force exertion on the environment by dynamically moving the \ac{com} closer to the front rotors during interactions.

\section{System Description}
\begin{figure}[!t]
   \centering
\includegraphics[trim={0.4cm 0.2cm 0.6cm 0.2cm},clip,width=\columnwidth]{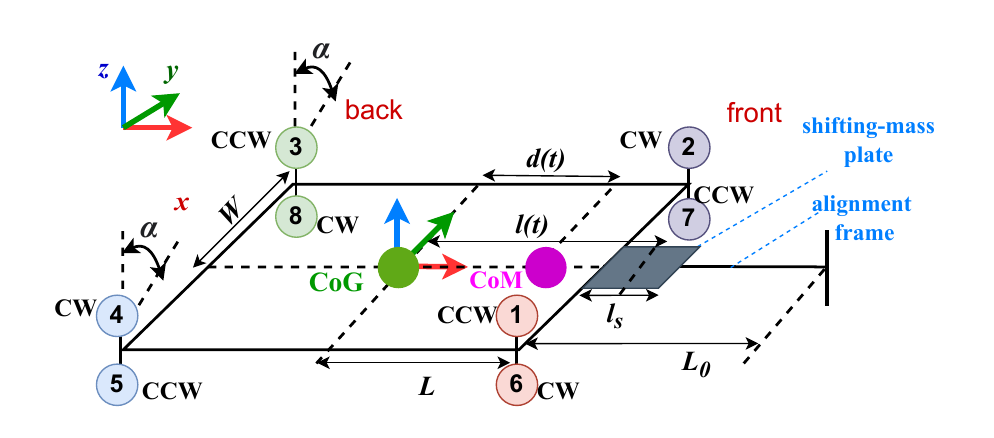}
    \caption{The world frame $\mathcal{F}_w=\{\bm{O};\bm{x},\bm{y},\bm{z}\}$ and the body frame $\mathcal{F}^B$. The rotors 3, 4, 5, and 8 can tilt simultaneously with $\alpha \in [-90\degree, \ 90\degree]$ around $\bm{y}_B$\new{;} the system's \ac{com} can shift along $\bm{x}_B$. CCW: counter-clockwise; CW: clockwise. The \ac{com} displacement $d(t)$ is resulted from moving the shifting-mass plate to the position $l(t)$ \ac{wrt} the body frame origin \cite{tmech}.}
    \label{fig:system}
    %\vspace{-0.5cm}
\end{figure}
In this section, we detail the system design of the aerial vehicle enabling pushing on vertical surfaces adapting the design concept introduced in Sec.~\ref{sec:design_concept}. 
\subsection{Overall System Design}
The proposed prototype adopts the form of an H-shaped coaxial octocopter, with the body frame $\mathcal{F}^B=\{O_B;\bm{x}_B,\bm{y}_B,\bm{z}_B\}$ attached to its \ac{cog}, as in Fig.~\ref{fig:physical_model}. The aerial vehicle's rotor configuration follows the order in Fig.~\ref{fig:system}. Rotors $1$, $2$, $6$, and $7$ in the front have fixed rotating axes parallel to $\bm{z}_B$. Rotors $3$, $4$, $5$, and $8$ on the back have parallel tiltable axes that can simultaneously tilt around $\bm{y}_B$ via a servo motor. We denote $\alpha \in [-90\degree, \ 90\degree]$ as the tilting angle between each tiltable axis and $\bm{z}_B$. 

The designed system features 5-\ac{dof} actuation enabling interactions with non-horizontal surfaces using a rigidly attached link, referred to as the \textit{alignment frame}, see Fig.~\ref{fig:physical_model}. The distance between the system \ac{cog} and each rotor center along $\bm{x}_B$ is represented by $L$, while the distance along $\bm{y}_B$ is denoted as $W$. $L_0$ is the length of the alignment frame.

We assume that the system's \ac{com} remains still during free flight and can displace along the body axis $\bm{x}_B$ during interactions. The \ac{com} displacing is accomplished by the linear motion of a movable plate equipped with major heavy components, including the battery, the onboard computer, the manipulator with heavy tools at the \ac{ee} tip, and other mechatronic components, referred to as the \textit{shifting-mass} \new{\textit{plate}}. 

Positioning the shifting-mass plate at the aerial vehicle's \ac{cog}, located within the rotor-defined area, allows us to assume that the system's initial \ac{com} coincides with the \ac{cog} during free flight.
The platform can interact with the environment along $\bm{x}_B$, denoted as the \textit{interaction axis}. The alignment frame at the front of the aerial vehicle outside the rotor-defined area preserves physical contact with the vertical surface while shifting-mass plate is maneuvered towards the work surface. 
\begin{figure}[!t]
   \centering
\begin{subfigure}{0.5\columnwidth}\centering
    \includegraphics[trim={0.6cm 0.2cm 0.6cm  0.2cm},clip,width=\columnwidth]{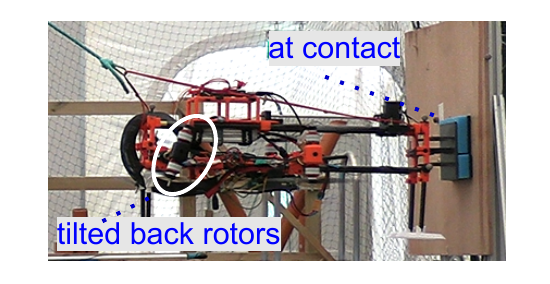}
    \caption{}
    \label{fig:action1}
\end{subfigure}
     \begin{subfigure}{0.48\columnwidth}\centering
    \includegraphics[trim={0.6cm 0.2cm 0.6cm  0.2cm},clip,width=\columnwidth]{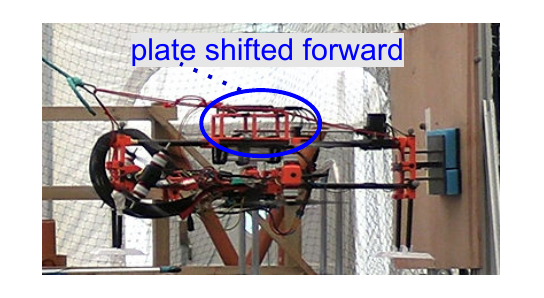}
    \caption{}
    \label{fig:action2}
\end{subfigure}
\caption{(a) The aerial vehicle establishes contact with the work surface using the alignment frame. (b) The shifting-mass plate is shifted closer to the work surface to have a displaced \ac{com} location closer to the front rotors for generating higher forces\cite{tmech}.}\label{fig:action_real}
    %\vspace{-0.5cm}
\end{figure}
\subsection{Center-of-Mass Displacing Feature}\label{sec:mass_shift}
We denote $m_S, \ m \in \mathbb{R}^+$ as the mass of the shifting-mass plate and the total system respectively, with $m_S<m$. The shifting-mass plate is connected to a linear actuator and can move along $\bm{x}_B$ starting from the system's \ac{cog} up to the alignment frame tip. 

The \ac{com} shifting process is enabled \new{only} when the system is preserving a stable contact with the work surface using the alignment frame tip, see Fig.~\ref{fig:action1} and \new{Fig.}~\ref{fig:action2}. We use the \ac{cog} of the shifting-mass plate to present its location. We denote $l_S$ as the length of the shifting-mass plate along the body axis $\bm{x}_B$, as in Fig.~\ref{fig:system}. The shifting-mass plate position along $\bm{x}_B$ \ac{wrt} the \new{body frame origin $\bm{O}_B$} is denoted as $l(t) \in [0,L+L_0-0.5l_S]$. By changing the shifting-mass plate position, the resultant displacement of the system's \ac{com} along $\bm{x}_B$ is denoted by $d(t)$. 

When the system's \ac{com} is located at $d>L$ outside the rotor-defined area (i.e., over-displaced), the aerial vehicle \new{often} flips around the contact area, leading to instability. Therefore, $d(t)$ is restricted by $[0, L]$ \new{to avoid risky scenarios and damage to the platform}. Assuming symmetric mass distribution \new{of the platform} around its body axes, the relation between $l(t)$ and $d(t)$ is given by: 
\begin{equation}\label{eq:shift_com}
    d(t)=\frac{m_S}{m}l(t).
\end{equation}
The maximum \ac{com} displacement $d=L$ results in a shifting-mass \new{plate} position of $l=\displaystyle\tfrac{m}{m_S}L>L$.
\section{Practical Challenges}
\label{sec:problem}
Ideally, it is desired to have the system's \ac{com} shifted to the maximum displacement $d=L$ during interactions achieving the optimal \ac{com} location $r_l=0$ in Fig.~\ref{fig:force_com}. Neglecting friction forces from the work surface, with $d=L$, only front rotors supply the force for gravity compensation. The thrust generated by back rotors can be fully used to exert horizontal interaction forces on the vertical surface with $\alpha=90 \degree$. In this case, the system can entail maximized horizontal force generation on vertical surfaces during pushing.

A physical prototype enabling simple pushing tasks using the alignment frame has been developed, as shown in Fig.~\ref{fig:action_real}. In practice, the mass distribution across the platform is not symmetric around the body axes, causing modeling inaccuracy. Directly moving the shifting-mass plate with $l=\displaystyle\frac{m}{m_S}L$ to achieve the optimal \ac{com} location ($r_l=0$ as introduced in Sec.~\ref{sec:mass_shift}) may still lead to over-displaced \ac{com} resulting in risky situations for the platform.

In reality, static friction forces often occur at contact assuming non-slipping conditions while pushing~\cite{tong_icra_2}. The static friction forces act as reaction forces from the environment and may assist in stabilizing the system even though the system's \ac{com} is slightly over-displaced. However, the maximum magnitude of the allowed static friction forces before slipping strictly corresponds to the materials of the \ac{ee} tip and the wall, often being difficult to predict and vary at different conditions. Therefore, blindly relying on the static friction forces to stabilize the system when the \ac{com} is over-displaced can lead to unpredictable crashes.

\section{Center-of-Mass Self-displacement}  
\label{sec:com_locate}
In this section, we propose a method enabling the platform to automatically position its shifting-mass plate to achieve the optimal \ac{com} displacement despite the external disturbances caused by friction forces and modeling inaccuracy identified in Sec.~\ref{sec:problem}.
\subsection{Flight Control Scheme}
\begin{figure}[!t]
   \centering
   %\vspace{0.25cm}
\includegraphics[trim={0.4cm 0.2cm 0.4cm 0.4cm},clip,width=\columnwidth]{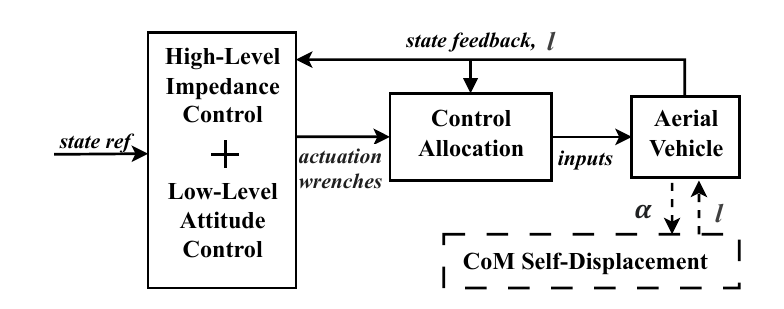}
    \caption{Flight control scheme of the aerial vehicle and the block illustrating the CoM self-displacement approach.}
    \label{fig:control_scheme}
    %\vspace{-0.5cm}
\end{figure}
A cascade control framework is implemented to handle both free flight and physical interaction tasks of the designed aerial vehicle, see Fig.~\ref{fig:control_scheme}. The control framework includes high-level impedance control, low-level attitude control, and control allocation. The high-level and the low-level controllers output desired actuation wrenches by tracking the linear and angular motion of the system, respectively. The control allocation computes system inputs including propeller speed and tilting axis angle from the desired actuation wrenches. We refer to \cite{ral,tmech} for the detailed control design. Pushing tasks are achieved by sending setpoints \enquote{inside} \cite{review} the work surface to the high-level impedance controller. The tracking performance during free flights and the pushing task execution using the proposed control design were validated via both Matlab\&Simulink-based simulations and physical experiments using the built prototype. Please note that external force estimation that enables accurate force control is not applied in the current settings considering the task requirements, but is under consideration for future developments.

\subsection{CoM Self-Displacement Scheme}
In addition to the developed controller,
in this section, we propose a self-displacing process to automatically find the \new{ideal} shifting-mass \new{plate} position \new{$l^*$} such that the system's \ac{com} reaches its maximum displacement $d^*=L$ at the optimal location introduced in Sec.~\ref{sec:optimal}. This process is executed while the platform stably pushes on a vertical surface using the alignment frame.

The \new{ideal} shifting-mass \new{plate} position \new{allows the maximum tilting value for the} back rotors\new{: $\alpha_{max}=90\degree$}. With the high-level impedance control, the platform can exert a force along the interaction axis $\bm{x}_B$ to the work surface with a setpoint position \enquote{inside} the work surface \cite{review,ral}. To achieve maximum tilting angle $\alpha$ during interactions, significant horizontal force generation on the vertical surface is required from the aerial system by setting a proper position reference. During the \new{self-displacing} process, while the system is exerting a horizontal force on the vertical surface using the alignment frame, the shifting-mass \new{plate} position is feedback controlled using the tilting angle $\alpha$ to reach $\alpha\new{_{max}}=90\degree$ and is given by:
\begin{equation}\label{eq:com_locate}
    \new{l=k_{\alpha} e_{\alpha}+k_{\alpha I} \int e_{\alpha}},
\end{equation}
where \new{$e_{\alpha}=\alpha_{max}-\alpha$} is the tracking error, $k_{\alpha}$ \new{and $k_{\alpha I}$ are positive gains.} 

Once the shifting-mass \new{plate} position reaches a steady state with $\alpha\approx90\degree$, we determine the ideal shifting-mass \new{plate} position $l^*$ for the current setup. This self-\new{displacing} process \new{essentially serves as a calibration step being} performed whenever there are changes in manipulators or other components affecting the system's mass distribution. As discussed in \new{Sec.}~\ref{sec:mass_shift}, considering the physical constraints imposed by the alignment frame's length, it is preferable to ensure $l^*<L+L_0-0.5l_S$ in hardware design to enable maximized \ac{com} displacement.

\section{Tool Manipulation}
To explore the advantages and functionality of the proposed aerial vehicle in advanced tool manipulation applications, we study the case where a 2-\ac{dof} manipulator is mounted on the shifting-mass plate (see Fig.~\ref{fig:physical_model}). The manipulation arm equipped with a tool at the \ac{ee} tip is used to execute a pushing task on a vertical surface, potentially serving for a drilling task. 
For aerial vehicles equipped with high-\ac{dof} manipulators, servo actuators are generally used in the manipulator. To reduce the reaction forces and torques induced by the motion of the manipulator, these servos are usually placed as close as possible to the \ac{com} of the aerial vehicle \cite{review}.

Traditional aerial system design often features a fixed \ac{com} located within the rotor-defined area. During interactions with a work surface, preventing rotors from contacting the surface is essential to avoid crashes. This is typically achieved by placing the \ac{ee} at a considerable distance from the rotors, resulting in an extended moment arm between the \ac{ee} tip and the aerial vehicle's \ac{com}. 
The load generated by the heavy tool's weight and interaction forces at the \ac{ee} tip due to the extended moment arm can cause damage to these servos owing to their technical constraints.
\begin{figure}[!t]
   \centering
     \begin{subfigure}{0.5\columnwidth}\centering
    \includegraphics[trim={0.6cm 0.4cm 0.6cm 0.6cm},clip,width=\columnwidth]{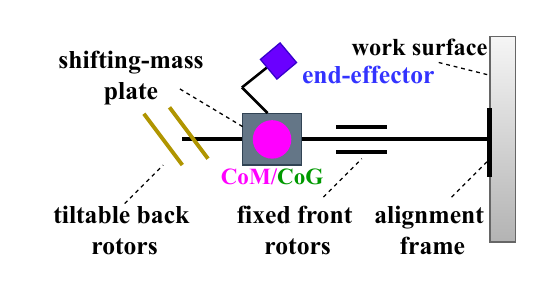}
    \caption{}
    \label{fig:bull_a}
\end{subfigure}
 \begin{subfigure}{0.45\columnwidth}\centering
    \includegraphics[width=0.9\columnwidth]{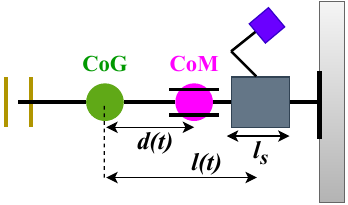}
    \caption{}
    \label{fig:bull_b}
\end{subfigure}\:\:\:
\begin{subfigure}{0.55\columnwidth}\centering
    \includegraphics[trim={0.6cm 0.4cm 0.4cm 0.4cm},clip,width=\columnwidth]{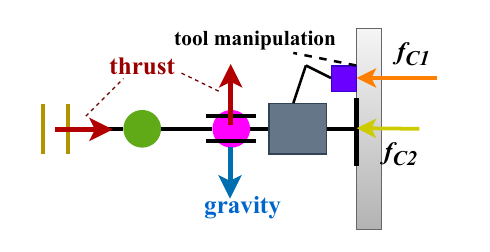}
    \caption{}
    \label{fig:bull_c}
\end{subfigure}
\begin{subfigure}{0.42\columnwidth}\centering
    \includegraphics[trim={0.9cm 0.6cm 0.9cm 0.6cm},clip,width=\columnwidth]{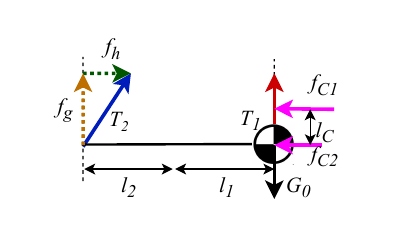}
    \caption{}
    \label{fig:fbd_2}
\end{subfigure}
\caption{(a-c) Schematics \new{illustrating} the \new{designed aerial vehicle's} operational flow for tool manipulation on a vertical surface\new{:}
(a) The alignment frame allows the platform to interact with the work surface.~(b) The shifting-mass plate changes its position $l(t)$ until the system \ac{com} reaches the maximum displacement at $d(t)=L$ while the back rotors can tilt up to $\alpha=90 \degree$.~(c) The manipulation task is executed closer to the work surface with the system \ac{com} aligned with the front rotors along the interaction axis. $f_{C2}$ and $f_{C1}$ are pushing forces acting on the alignment frame tip and the \ac{ee} tip, respectively. (d) Free body diagram of (c).}\label{fig:aero_bull}
    %\vspace{-0.5cm}
\end{figure}
\subsection{Ideal Operational Flow}\label{sec:operation}
We present the desired operational flow of the pushing task using the proposed aerial vehicle with an additional manipulator attached to the shifting-mass plate, as in Fig.~\ref{fig:aero_bull}.
With the proposed aerial vehicle, the manipulation arm stays close to the system's \ac{cog} during free flight, as in Fig.~\ref{fig:bull_a}. After establishing a stable contact with the vertical surface using the alignment frame, the manipulation arm shifts along with the shifting-mass plate toward the work surface as in Fig.~\ref{fig:bull_b}. With the \ac{com} self-displacement process introduced in Sec.~\ref{sec:com_locate}, we identify the optimal shifting-mass plate position $l^*$ of the corresponding setup. When the system's \ac{com} achieves its maximum displacement aligning with the front rotors, the manipulation arm executes the pushing operation on the work surface, as in Fig.~\ref{fig:bull_c}.

When at the optimal \ac{com} location, the system allows the shortest moment arm between the contact point and the \ac{com}. Such configuration enables the use of shorter beams for the manipulation arm design, leading to less torque load on the joints of the manipulation arm when exerting the same amount of force on the work surface. 

\subsection{Control Approach}
The main objective is to explore the benefits and feasibility of the designed aerial vehicle in tool manipulation. Therefore, we aim to minimize the control investigation during this preliminary study.
To execute the described task in Fig.~\ref{fig:aero_bull}, a decentralized control approach is applied to the system, in the sense that the aerial vehicle and the manipulation arm are controlled separately \cite{review_0}. The control of the aerial vehicle follows the scheme in Fig.~\ref{fig:control_scheme}. The position control of the 2-DoF robotic arm is well-established in the literature and out of the scope of this work. The movements of the robotic arm and the external wrenches caused by the interactions between the \ac{ee} tip and the work surface are considered perturbations for the aerial vehicle. Considering the operation flow described in the previous section, while the robotic arm is pushing on the work surface, the alignment frame also preserves contact with the work surface exerting an amount of force.
To ensure the validity of such a control strategy, the following practical limitations are considered. Operation guidelines are derived accordingly to guarantee successful task execution.
\subsection{Practical Limitations}
We assume negligible friction forces as in Sec.~\ref{sec:background} and define the pushing forces exerted from the alignment frame and the \ac{ee} tip are $f_{C2}$ and $f_{C1} \in \mathbb{R}$, as in Fig.~\ref{fig:bull_c}. 

Assuming lightweight manipulator design, we neglect the effects caused by the manipulator movement during tool operation. During the tool manipulation in Fig.~\ref{fig:bull_c}, the attitude controller maintains the aerial vehicle orientation in the presence of the disturbance caused by the external torque generated by $f_{C1}$ \ac{wrt} the \ac{com}, considering that $f_{C2}$ passes through the \ac{com}.  

Despite its effect on the attitude controller, it is also essential to understand the influence of this torque on the system's force exertion. To do so, we study the major forces and torques acting on the whole system during operation similar to Fig.~\ref{fig:design2} for the seek of simplicity. Assuming equilibrium while pushing, the simplified planar free body diagram of the system is depicted in Fig.~\ref{fig:fbd_2}. The following force relationships hold:
\begin{align}
    f_g&=\frac{f_{C1}l_C}{l_1+l_2}\label{eq:force_1},\\
f_{C1}+f_{C2}&=f_h=\sqrt{T_2^2-f_g^2}\label{eq:force_2}.
\end{align}
\begin{figure}[!t]
   \centering
   %\vspace{0.25cm}
\includegraphics[trim={0.4cm 0.4cm 0.4cm 0.4cm},clip,width=\columnwidth]{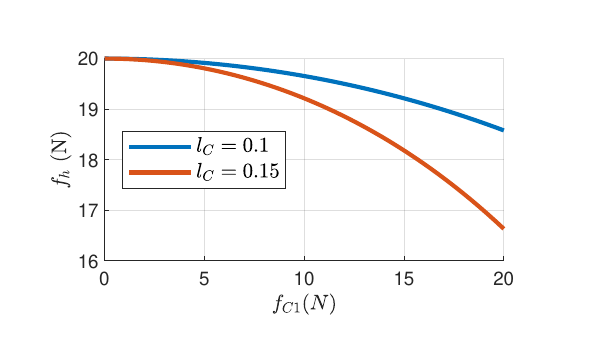}
    \caption{Force variation of $f_h$ along with increased $f_{C1}$ for $l_C=0.1, 0.15\si{\meter}$ based on \cref{eq:force_2}.}
    \label{fig:force_2}
    %\vspace{-0.5cm}
\end{figure}

When $l_C=0$, the force component $f_g$ of back rotor thrust $T_2$ opposing gravity is zero according to \cref{eq:force_1}, and the back rotor thrust can be fully utilized for exerting pushing forces as the ideal case shown in Fig.~\ref{fig:bull_c}. For $l_C\neq0$, a non-zero $f_g$ is demanded to compensate the external torque caused by $f_{C1}$, therefore, the back rotor thrust cannot tilt up to $\alpha=90\degree$ as expected. We analyze a case where $T_2=20\si{\newton}$ and $l_1+l_2=0.27\si{\meter}$ like in Sec.~\ref{sec:background}. The force variation of $f_h$ exerted from the system along with increased $f_{C1}$ and $l_C=0.1, \ 0.15\si{\meter}$ is depicted in Fig.~\ref{fig:force_2}. As the pushing force from the robotic arm increases, the allowed pushing force from the platform reduces. Moreover, the larger $l_C$ is, the smaller the allowed force exertion is.

\subsection{Operation Guidelines}
To achieve superior force exertion from the proposed aerial vehicle and ensure the adopted decentralized control approach, it is desired to minimize the external torque caused by the tool operation. This is achieved by selecting the operation point close to the system \ac{com} assumed to coincide with the alignment frame contact area. In the following section, we design the simulation scenario using an operation point with $l_C=0.05\si{\meter}$ considering the limited workspace of a 2-\ac{dof} manipulation arm and ensuring effective force generation from the platform.
\section{\new{Numerical Results}}\label{sec:validation}
In this section, simulations that closely emulate the physical platform displayed in \new{Fig.}~\ref{fig:action_real} are performed to test the proposed \ac{com} self-displacing process and highlight the advantage of the proposed system in tool manipulation with a 2-\ac{dof} robotic arm. The developed simulation environment fundamentally supported the physical experiments in our previous work with validated accuracy and reliability. These testing results therefore offer valuable guidelines and insights for future development. 
\subsection{Simulation Setup}
\new{We use} a Simscape model, as in \new{Fig.}~\ref{fig:physical_model}, \new{while the} controller developed in \cite{ral,tmech} and the \ac{com} self-\new{displacing} approach in \new{Sec.}~\ref{sec:com_locate} are implemented in Simulink \new{ running at 100 Hz}. \new{A 2-\ac{dof} manipulation arm equipped with a heavy tool at the \ac{ee} is mounted on the shifting-mass plate, potentially serving for a drilling task. The total system mass and the shifting mass used in the simulation are the same as the physical platform, where $m=3.12$\si{\kilogram} and $m_S=0.9$\si{\kilogram}. The simplified schematic of the manipulator is shown in Fig.~\ref{fig:joints}. We define $L_1$, $L_2 \in \mathbb{R}^+$ as the length of two beams, and $\tau_1$, $\tau_2 \in \mathbb{R}^+$ as the magnitude of joint torques.} The tests include two physical interaction tasks.

In \textbf{task 1}, the platform pushes on a vertical work surface using the alignment frame, and the \ac{com} self-displacing approach is conducted to find the ideal shifting-mass plate position for the current setup. In this task, the robotic arm remains still without operating on the surface or any other movements. In \textbf{task 2}, the manipulator is used to push on the vertical work surface. To highlight the advantage of our proposed system with a dynamic \ac{com} design, we conducted this task in two scenarios. In scenario (a), the shifting-mass plate with the robotic arm attached is shifted to its ideal position identified by \textbf{Task 1}, and the robotic arm operates close to the work surface. This test is illustrated in the accompanying video. In scenario (b), the shifting-mass plate remains still and the attached robotic arm operates further from the work surface compared to scenario (a).

\new{\subsection{\textbf{Task 1}}}
\new{This section presents the testing results for a pushing task on a vertical surface using only the alignment frame, validating the feasibility of the proposed \ac{com} self-displacing approach.} The static coefficient between the \ac{ee} tip and the wall is set to $0.5$. A set of way-points \new{culminating with} points \enquote{inside} \cite{ral,review} the work surface was sent to the controller to approach and \new{establish a }contact with the work surface using the alignment frame. Once stable contact between the system and the work surface was obtained, the \new{\ac{com} self-displacing process} was enabled. The shifting-mass \new{plate} position $l$ changed along the \new{positive} body axis $\bm{x}_B$ starting from its initial value $l=0$\si{\meter} while the back rotors tilting angle $\alpha$ was increasing, see \new{Fig.}~\ref{fig:interaction} Zone~1.

In Zone~2, the \new{\ac{com} self-displacing} was completed with $\alpha\approx90\degree$\new{:} the shifting-mass \new{plate} reached a position at $l=0.25$\si{\meter} and kept increasing until the \new{ideal} position at $l^*=0.281$\si{\meter}. It maintained this position while exerting horizontal forces on the vertical surface using the full thrust of back rotors. Afterward, the shifting-mass \new{plate} returned to its initial position while the tilting angle $\alpha$ was decreasing, as in Zone~3. \new{The determined \new{ideal} shifting-mass \new{plate} position $l^*$ from the simulation can assist in setting the maximum allowed shifting-mass plate position in the physical experiments to prevent instability due to \ac{com} over-displacement.}

The gain values $k_{\alpha}$ and $k_{\alpha I}$ are pivotal and need adjustment when there are changes in mass distribution. \new{In practice, having over-displaced \ac{com} (i.e. $d>L$) is highly risky. Therefore, smaller gain values are suggested at the beginning of the gain tuning process with the real platform to prevent damage. In the future, we will look into safely identifying the maximum \ac{com} displacement with the physical platform.} 

\subsubsection{Friction Effects}
\begin{figure}[!t]
   \centering
\includegraphics[trim={0.15cm 0.1cm 1.9cm 0.4cm},clip,width=\columnwidth]{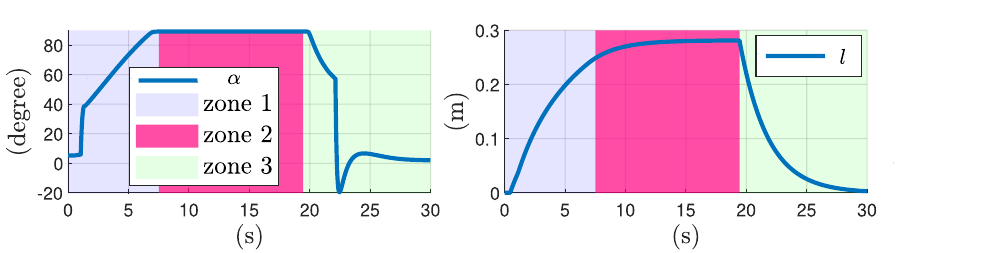}
    \caption{\new{\textbf{Task 1}:} the tilting angle $\alpha$ \new{(on the left)} and shifting-mass \new{plate} position $l$ \new{(on the right) evolution during the interaction phase}.}
    \label{fig:interaction}
\end{figure}

From \new{Fig.}~\ref{fig:interaction}, a range of $l\in[0.25 \ 0.281]$\si{\meter} can enable $\alpha\approx90 \degree$ instead of a single value of $l^*$. The friction forces at contact allowed greater \ac{com} displacement as expected. In the future, we aim to investigate the effects of different friction conditions on the \ac{com} displacement, leveraging friction forces to save energy.
\begin{figure}[!t]
   \centering
     \begin{subfigure}{0.2\columnwidth}
     \centering
    \includegraphics[height=2.5cm]{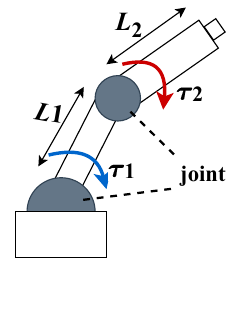}
    \caption{}
    \label{fig:joints}
\end{subfigure}
\begin{subfigure}{0.78\columnwidth}\centering
   \includegraphics[height=3cm]{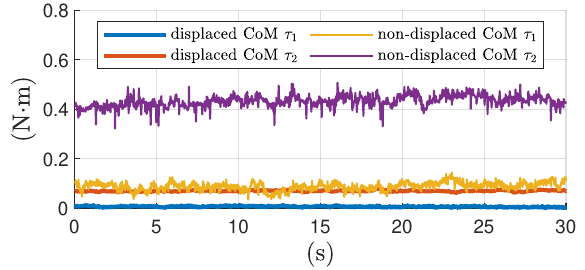}
    \caption{}
    \label{fig:joint_torques}
\end{subfigure}
\caption{\new{(a) A simplified schematic of the 2-\ac{dof} manipulator. (b) \textbf{Task 2}: the 2-\ac{dof} manipulator joint torque magnitude $\tau_1$ and $\tau_2$ in two different scenarios: displaced \ac{com} and non-displaced \ac{com}.}}
\label{fig:benchmark}
    %\vspace{-0.5cm}
\end{figure}
\new{\subsection{\textbf{Task 2}}}
\new{In this task, the attached 2-\ac{dof} manipulator was used to push on the work surface under two scenarios: scenario (a) - displaced \ac{com}, and scenario (b) - non-displaced \ac{com}. In scenario (a), the shifting-mass plate equipped with the manipulator was shifted to its ideal position $l^*=0.28$\si{\meter} identified in \textbf{Task 1}. Operating close to the work surface with the maximum \ac{com} displacement allowed shorter beams of the manipulator. In this scenario, the manipulator was designed with $L_1=L_2=0.05$\si{\meter}. In scenario (b), the shifting-mass plate maintained its initial position, i.e. there was no \ac{com} displacement (the change of the system \ac{com} due to different arm configuration was not considered). In this case, longer beams were used for the manipulator with $L_1=L_2=0.2$\si{\meter} which are three times longer than the ones in scenario (a).}

In both scenarios, the manipulator was required to exert a $5$\si{\newton} pushing force normal to the work surface, for the manipulator control we refer to \cite{siciliano}. The joint torque magnitude of the manipulator for both scenarios is displayed in Fig.~\ref{fig:joint_torques}. In scenario (b), the joint torque magnitude reached $0.4$\si{\newton\meter} which is significantly larger than the magnitude in scenario (a), as expected in Sec.~\ref{sec:operation}. With the simulation results, we validated the proposed \ac{com} self-displacing approach to enhance the force exertion of the platform. Moreover, the operation flow in Fig.~\ref{fig:aero_bull} is demonstrated in the simulation emphasizing the functionality and benefits of the proposed aerial system in advanced tool manipulation.

\section{Conclusion}\label{sec:conclusion}
Building on our previous research, where we introduced a novel aerial vehicle with a dynamic \ac{com} configuration, we aim to enhance the system's manipulation capabilities in this work. To address this, we proposed a novel method for displacing the system's \ac{com} to its optimal location maximizing the platform's force generation during physical interactions in the presence of modeling inaccuracy and friction effects. Moreover, by integrating a 2-\ac{dof} manipulation arm to the presented aerial vehicle, we have expanded the functionality of these aerial manipulators enabling more advanced tool manipulation tasks. We identified the practical limits of such operations through in-depth analysis and provided operation guidelines to achieve superior behavior during task execution with the proposed aerial vehicle and control approach.  Our simulations confirm the effectiveness of these innovations, highlighting their potential to transform practices. This work opens up promising future venues for employing this new class of aerial vehicles with varied \ac{com} locations in advanced aerial manipulation.

\addtolength{\textheight}{-12cm}   % This command serves to balance the column lengths

%%%%%%%%%%%%%%%%%%%%%%%%%%%%%%%%%%%%%%%%%%%%%%%%%%%%%%%%%%%%%%%%%%%%%%%%%%%%%%%%

%%%%%%%%%%%%%%%%%%%%%%%%%%%%%%%%%%%%%%%%%%%%%%%%%%%%%%%%%%%%%%%%%%%%%%%%%%%%%%%%



% \section*{ACKNOWLEDGMENT}

% ...

%%%%%%%%%%%%%%%%%%%%%%%%%%%%%%%%%%%%%%%%%%%%%%%%%%%%%%%%%%%%%%%%%%%%%%%%%%%%%%%%

\begin{thebibliography}{99}
\bibitem{review} A. Ollero, M. Tognon, A. Suarez, D. Lee and A. Franchi, "Past, Present, and Future of Aerial Robotic Manipulators," in IEEE Transactions on Robotics, vol. 38, no. 1, pp. 626-645, Feb. 2022, doi: 10.1109/TRO.2021.3084395. 

\bibitem{tong_iser} T. Hui, F. Braun, N. Scheidt, M. Fehr, and M. Fumagalli, Versa-
tile airborne ultrasonic ndt technologies via active omni-sliding
with over-actuated aerial vehicles, in: M. H. Ang Jr, O. Khatib
(Eds.), Experimental Robotics, Springer Nature Switzerland,
Cham, 2024, pp. 450–460. doi:https://doi.org/10.1007/
978-3-031-63596-0-40.

\bibitem{bart} T. Bartelds, A. Capra, S. Hamaza, S. Stramigioli and M. Fumagalli, "Compliant Aerial Manipulators: Toward a New Generation of Aerial Robotic Workers," in IEEE Robotics and Automation Letters, vol. 1, no. 1, pp. 477-483, Jan. 2016, doi: 10.1109/LRA.2016.2519948.

\bibitem{truj} Trujillo, M.Á.; Martínez-de Dios, J.R.; Martín, C.; Viguria, A.; Ollero, A. Novel Aerial Manipulator for Accurate and Robust Industrial NDT Contact Inspection: A New Tool for the Oil and Gas Inspection Industry. Sensors 2019, 19, 1305. https://doi.org/10.3390/s19061305

\bibitem{watson2022} R. Watson et al., "Dry Coupled Ultrasonic Non-Destructive Evaluation Using an Over-Actuated Unmanned Aerial Vehicle," in IEEE Transactions on Automation Science and Engineering, vol. 19, no. 4, pp. 2874-2889, Oct. 2022, doi: \url{10.1109/TASE.2021.3094966}.

\bibitem{bodie2019}K. Bodie et al., “An omnidirectional aerial manipulation platform for
contact-based inspection,” in Proc. Robot.: Sci. Syst., 2019.

\bibitem{tong_icra_1} T. Hui, M. J. Fernandez Gonzalez, and M. Fumagalli, "Safety-Conscious Pushing on Diverse Oriented Surfaces with Underactuated Aerial Vehicles," 2024 International Conference on Robotics and Automation (ICRA), Yokohama, Japan, 2024, doi: arXiv:2402.15243.

\bibitem{tong_icra_2} T. Hui, E. Cuniato, M. Pantic, M. Tognon,M. Fumagalli, and R. Siegwart, "Passive Aligning Physical Interaction of Fully-Actuated Aerial Vehicles
for Pushing Tasks," 2024 International Conference on Robotics and Automation (ICRA), Yokohama, Japan, 2024, doi:  arXiv:2402.17434v1.

\bibitem{arm} A. Suarez, A. Gonzalez-Morgado and A. Ollero, "Lightweight and Compliant Bilateral Teleoperation System with Anthropomorphic Arms for Aerial and Ground Service Operations," 2024 IEEE International Conference on Robotics and Automation (ICRA), Yokohama, Japan, 2024, pp. 15685-15691, doi: 10.1109/ICRA57147.2024.10611383.

\bibitem{sun} Y. Sun, A. Plowcha, M. Nail, S. Elbaum, B. Terry and C. Detweiler, "Unmanned Aerial Auger for Underground Sensor Installation," 2018 IEEE/RSJ International Conference on Intelligent Robots and Systems (IROS), Madrid, Spain, 2018, pp. 1374-1381, doi: 10.1109/IROS.2018.8593824. 

\bibitem{drill} R. Dautzenberg et al., "A Perching and Tilting Aerial Robot for Precise and Versatile Power Tool Work on Vertical Walls," 2023 IEEE/RSJ International Conference on Intelligent Robots and Systems (IROS), Detroit, MI, USA, 2023, pp. 1094-1101, doi: 10.1109/IROS55552.2023.10342274.

\bibitem{ding} C. Ding and L. Lu, "A Tilting-Rotor Unmanned Aerial Vehicle for Enhanced Aerial Locomotion and Manipulation Capabilities: Design, Control, and Applications," in IEEE/ASME Transactions on Mechatronics, vol. 26, no. 4, pp. 2237-2248, Aug. 2021, doi: 10.1109/TMECH.2020.3036346.

\bibitem{mina} M. Kamel et al., "The Voliro Omniorientational Hexacopter: An Agile and Maneuverable Tiltable-Rotor Aerial Vehicle," in IEEE Robotics \& Automation Magazine, vol. 25, no. 4, pp. 34-44, Dec. 2018, doi: 10.1109/MRA.2018.2866758.

\bibitem{antonio} M. Ryll, G. Muscio, F. Pierri, E. Cataldi, G. Antonelli, F. Cac-
cavale, D. Bicego, A. Franchi, 6d interaction control with
aerial robots: The flying end-effector paradigm, The Inter-
national Journal of Robotics Research 38 (2019) 1045–1062. doi:10.1177/0278364919856694.

\bibitem{odar} S. Park, J. Lee, J. Ahn, M. Kim, J. Her, G.-H. Yang, D. Lee,
Odar: Aerial manipulation platform enabling omnidirectional
wrench generation, IEEE/ASME Transactions on Mechatronics
23 (2018) 1907–1918. doi:10.1109/TMECH.2018.2848255.

\bibitem{hwang} S. Hwang, D. Lee, C. Kim and H. J. Kim, "Autonomous Heavy Object Pushing Using a Coaxial Tiltrotor," in IEEE Transactions on Automation Science and Engineering, doi: 10.1109/TASE.2024.3409058.

\bibitem{meng} J. Meng, J. Buzzatto,Y. Liu, M. Liarokapis. On Aerial Robots with Grasping and Perching Capabilities: A Comprehensive Review. Front Robot AI. 2022 Mar 25;8:739173. doi: 10.3389/frobt.2021.739173. 

\bibitem{backus} S. B. Backus, L. U. Odhner and A. M. Dollar, "Design of hands for aerial manipulation: Actuator number and routing for grasping and perching," 2014 IEEE/RSJ International Conference on Intelligent Robots and Systems, Chicago, IL, USA, 2014, pp. 34-40, doi: 10.1109/IROS.2014.6942537. 

\bibitem{zhang} H. Zhang, J. Sun and J. Zhao, "Compliant Bistable Gripper for Aerial Perching and Grasping," 2019 International Conference on Robotics and Automation (ICRA), Montreal, QC, Canada, 2019, pp. 1248-1253, doi: 10.1109/ICRA.2019.8793936. 

\bibitem{daler} L. Daler, A. Klaptocz, A. Briod, M. Sitti and D. Floreano, "A perching mechanism for flying robots using a fibre-based adhesive," 2013 IEEE International Conference on Robotics and Automation, Karlsruhe, Germany, 2013, pp. 4433-4438, doi: 10.1109/ICRA.2013.6631206. 


\bibitem{wop} H. W. Wopereis, T. D. van der Molen, T. H. Post, S. Stramigioli and M. Fumagalli, "Mechanism for perching on smooth surfaces using aerial impacts," 2016 IEEE International Symposium on Safety, Security, and Rescue Robotics (SSRR), Lausanne, Switzerland, 2016, pp. 154-159, doi: 10.1109/SSRR.2016.7784292. 


\bibitem{ral} T. Hui, S. Rucareanu, E. Zamora, S. D'Angelo, M. Fumagalli, "Dynamic Center-of-Mass Displacement in Aerial Manipulation: An Innovative Platform Design," 2024, https://doi.org/10.48550/arXiv.2404.01110.

\bibitem{tmech} T. Hui, E. Zamora, S. D'Angelo, S. Rucareanu, M. Fumagalli, "AEROBULL: A Center-of-Mass Displacing Aerial Vehicle Enabling Efficient High-Force Interaction," 2024, 
https://doi.org/10.48550/arXiv.2408.15008.

\bibitem{lee2013} F. Goodarzi, D. Lee and T. Lee, "Geometric nonlinear PID control of a quadrotor UAV on SE(3)," 2013 European Control Conference (ECC), Zurich, Switzerland, 2013, pp. 3845-3850, doi: 10.23919/ECC.2013.6669644. 

\bibitem{siciliano} B. Siciliano, L. Sciavicco, L. Villani, and G. Oriolo. 2010. Robotics: Modelling, Planning and Control. Springer Publishing Company, Incorporated.

\bibitem{review_0} F. Ruggiero, V. Lippiello and A. Ollero, "Aerial Manipulation: A Literature Review," in IEEE Robotics and Automation Letters, vol. 3, no. 3, pp. 1957-1964, July 2018, doi: 10.1109/LRA.2018.2808541.

% \bibitem{kim} S. Kim, S. Choi and H. J. Kim, "Aerial manipulation using a quadrotor with a two DOF robotic arm," 2013 IEEE/RSJ International Conference on Intelligent Robots and Systems, Tokyo, Japan, 2013, pp. 4990-4995, doi: 10.1109/IROS.2013.6697077. 

% \bibitem{park} H. -N. Nguyen, S. Park, J. Park and D. Lee, "A Novel Robotic Platform for Aerial Manipulation Using Quadrotors as Rotating Thrust Generators," in IEEE Transactions on Robotics, vol. 34, no. 2, pp. 353-369, April 2018, doi: 10.1109/TRO.2018.2791604.

% \bibitem{fabio} F. Ruggiero et al., "A multilayer control for multirotor UAVs equipped with a servo robot arm," 2015 IEEE International Conference on Robotics and Automation (ICRA), Seattle, WA, USA, 2015, pp. 4014-4020, doi: 10.1109/ICRA.2015.7139760. 

% \bibitem{grinding} M.A. Elbestawi, K.M. Yuen, A.K. Srivastava, H. Dai, Adaptive Force Control for Robotic Disk Grinding, CIRP Annals, Volume 40, Issue 1, 1991, Pages 391-394, ISSN 0007-8506, https://doi.org/10.1016/S0007-8506(07)62014-9.

% \bibitem{tong_aim} T. Hui and M. Fumagalli, "Static-Equilibrium Oriented Interaction Force Modeling and Control of Aerial Manipulation with Uni-Directional Thrust Multirotors," 2023 IEEE/ASME International Conference on Advanced Intelligent Mechatronics (AIM), Seattle, WA, USA, 2023, pp. 452-459, doi: 10.1109/AIM46323.2023.10196117.





% \bibitem{laren} A. McLaren, Z. Fitzgerald, G. Gao and M. Liarokapis, "A Passive Closing, Tendon Driven, Adaptive Robot Hand for Ultra-Fast, Aerial Grasping and Perching," 2019 IEEE/RSJ International Conference on Intelligent Robots and Systems (IROS), Macau, China, 2019, pp. 5602-5607, doi: 10.1109/IROS40897.2019.8968076.



% \bibitem{lee2013} F. Goodarzi, D. Lee and T. Lee, "Geometric nonlinear PID control of a quadrotor UAV on SE(3)," 2013 European Control Conference (ECC), Zurich, Switzerland, 2013, pp. 3845-3850, doi: 10.23919/ECC.2013.6669644. 




% \bibitem{dangelo} S. D'Angelo et al., "Stabilization and control on a pipe-rack of a wheeled mobile manipulator with a snake-like arm" in Robotics and Autonomous Systems, vol. 171, Jan. 2024, doi: https://doi.org/10.1016/j.robot.2023.104554.

% \bibitem{aero_tong} T. Hui et al., "Centroidal Aerodynamic Modeling and Control of Flying Multibody Robots," 2022 International Conference on Robotics and Automation (ICRA), Philadelphia, PA, USA, 2022, pp. 2017-2023, doi: 10.1109/ICRA46639.2022.9812147. 

% \bibitem{siciliano} B. Siciliano, L. Sciavicco, L. Villani, and G. Oriolo. 2010. Robotics: Modelling, Planning and Control. Springer Publishing Company, Incorporated.


\end{thebibliography}
\end{document}